\RequirePackage{fix-cm}
\documentclass[twocolumn]{svjour3}          
\smartqed  
\usepackage{graphicx}
\begin{document}

\title{Deep Learning Based Traffic Surveillance System For Missing and Suspicious Car Detection
}
\subtitle{}


\author{K.V. Kadambari*       \and
        Vishnu Vardhan Nimmalapudi
}


\institute{*Corresponding author: K.V. Kadambari \at Assistant Professor, Dept. of Computer Science and Engg, National Institute of Technology, Warangal, Telangana, India\\
              \email{kadambari@nitw.ac.in}           
           \and
           Vishnu Vardhan Nimmalapudi \at B.Tech., Dept. of Electronics and Communication Engg,National Institute of Technology, Warangal, Telangana, India\\
           \email{nvishnu@student.nitw.ac.in}\\
           }

\date{Received:  / Accepted: }

\maketitle

\begin{abstract}
Vehicle theft is arguably one of the fastest-growing types of crime in India. In some of the urban areas, vehicle theft cases are believed to be around 100 each day.  Identification of stolen vehicles in such precarious scenarios is not possible using traditional methods like manual checking and radio frequency identification(RFID) based technologies. This paper presents a deep learning based automatic traffic surveillance system for the detection of stolen/suspicious cars from the  closed circuit television(CCTV) camera footage. It mainly comprises of four parts: Select-Detector, Image Quality Enhancer, Image Transformer, and Smart Recognizer. The Select-Detector is used for extracting the frames containing vehicles and to detect the license plates much efficiently with minimum time complexity. The quality of the license plates is then enhanced using Image Quality Enhancer which uses pix2pix generative adversarial network(GAN) for enhancing the license plates that are affected by temporal changes like low light, shadow, etc. Image Transformer is used to tackle the problem of inefficient recognition of license plates which are not horizontal(which are at an angle) by transforming the license plate to different levels of rotation and cropping. Smart Recognizer recognizes the license plate number using Tesseract optical character recognition(OCR) and corrects the wrongly recognized characters using Error-Detector. The effectiveness of the proposed approach is tested on the government's CCTV camera footage, which resulted in identifying the stolen/suspicious cars with an accuracy of 87\%.
\end{abstract}

\section{Introduction}
\label{intro}

The number of vehicles have increased considerably in the past few decades and the number of stolen vehicle count and their uses in crime activities is on the rise too. According to the police department in Delhi,India, vehicle theft is one of the least-solvable offenses. In 2018 alone, over 44,000 vehicles were stolen but less than 20\% of them were recovered.
Traditional methods like inspection of the vehicle by halting the traffic for finding these cars have become obsolete. It is important to make use of technology to identify suspicious/stolen vehicles. It is difficult to manually search each and every car from the CCTV footage. So there is a need for a system that can automatically track the location and time of the suspicious car within a city once provided with the license plate number.

In most of the developed countries, Red Light Cameras\cite{b1} are used and license plates are strictly maintained. But in India, it’s only confined to CCTV cameras and the license plates lack standardization. The problem with CCTV cameras is the low resolution which results in poor recording. The attributes to consider standardization are size, font, color of the plate, the distance between two characters etc. The complexity of recognizing a license plate becomes even harder in India as the license plates often come with regional fonts. Hence, the system needs to be trained for each region.

Automation of the process of tracking suspicious cars is highly desirable as it signiﬁcantly reduces the number of human resources needed. Over the recent years, numerous technologies consisting of innovative methods to extract vehicle number plate have emerged but it yet proves to be a difficult task. Major challenges are due to low-resolution CCTV footages and due to the temporal changes in conditions like illumination, shadows, low light, bad weather etc.

In recent times, there has been a lot of research done on license plate recognition recently \cite{b2,b3,b4,b5,b6,b7,b8}.
Saini et al.\cite{b3}  designed a deep learning based system where  K-Nearest Neighbors algorithm and convolutional neural network classifier are used to identify the stolen/suspicious vehicles without human intervention. In the first stage, Google's  Tensor Flow detection API  is used to detect the vehicle with sub-modules to identify registration number and color from a real-time video stream. In the second stage, the extracted characteristic i.e. vehicle registration number and color are compared with the Regional Transport Office(RTO) record.Their model is tested on a database containing real time road traffic videos captured with camera. This design is not able to tackle the challenges like identification of license plates when they are not horizontal(i.e at an angle) and in conditions of low light, shadow, blurrness of license plates etc. 

Seungwon et al.\cite{b2} proposed an integrated video-based automobile tracking system(IVATS), based on Kafka and HBase. This system consists of Frame Distributor for distributing the video frames from video sources like car dashboard cameras, drone~-mounted, CCTV cameras.It uses a Feature Extractor for extracting principal vehicle features such as location, plate number, and time from each frame, and an Information Manager for storing all features into a database and retrieving them for query processing.
Tesseract-OCR is used for license plate recognition. This approach may work satisfactorily for tracking some vehicles. However, their Feature Extractor can't be taken as an ultimate solution for identifying license plates for the same reasons mentioned above like plate at an angle,low light etc. The failure of Tesseract-OCR in recognizing such license plates and the solution to this problem, are addressed in the later sections.

Sandesh\cite{b9} proposed a track by detection framework to track the moving vehicles on the road. You Only Look Once(YOLO)v3 object detection system was used to detect the vehicles and the concepts of Deep Simple Online and Realtime Tracking(SORT) algorithm were applied for tracking. In tracking by detection, each frame uses an object detector to find possible instances of objects and then matches those detections with corresponding objects in the preceding frame. This tracking is purely based on a particular car detection over a period of time. The major drawback of this system is that it can’t track the vehicles based on the license plates. It needs a car image to track it, in the given set of videos.

Sarfraz et al.\cite{b10} proposed a framework for real-time automatic license plate recognition, comprising of detection/localization, tracking, and recognition of license plates in CCTV surveillance videos. Firstly, the license plate is localized in the incoming frames using background learning and then using histogram of oriented gradients(HOG) on the region of selection. Tracking of the located plate is further done in each frame by continuous upgradation of background and ﬁnding the new location of the license plate. On the detected plate(s), the character recognition procedure is applied to recognize the characters of the license plate in each frame using the nearest neighbor classiﬁer.The framework is evaluated on a set of CCTV road surveillance videos obtained for general purpose and manual inspection. The disadvantage of this method is that it fails to recognize the plates at an angle.

To make a full-fledged working system, it should be able to tackle the following challenges like recognizing license plates which are at an angle as most of the CCTV cameras are not fixed always at the center of the road, the videos may contain vehicles that are not exactly parallel or horizontal to the camera view.  There is also a need for enhancing the license plate quality as the videos may be taken in different lighting conditions. The complexity of the frames extraction process can also be decreased by using a proper algorithm which can extract only those frames that contain vehicles, instead of extracting every frame from the CCTV videos. The currently available systems for license plate detection fail to tackle the above-mentioned challenges. However, the novelty of the proposed framework is in its ability to tackle the above-mentioned challenges and to detect suspicious cars with better accuracy using their license plates.

The layout of the rest of the paper is as follows. Section II delineates the proposed framework of the paper. Section III contains the analysis of our experiments and results. Section IV illustrates the conclusion of the experiment and discusses future areas of exploration.

\section{Proposed Framework}
\label{sec:1}
This section describes the proposed framework for the suspicious/missing car identification using the traffic surveillance videos according to Fig. 1. There are four stages: Select-Detector, Image Quality Enhancer, Image Transformer, Smart Recognizer. The Select-Detector performs two functions. First is extracting of the frames, which contain vehicles, second is to detect the location of the license plate in the extracted frames using YOLOv3\cite{b11} framework.YOLO is a state-of-the-art, real-time object detection system, which is one of the most effective object detection algorithms. It is extremely fast and accurate. It is a convolutional neural network that divides the image into regions and predicts bounding boxes and probabilities for each region. These bounding boxes are weighted by the predicted probabilities. YOLOv3 is the faster vesion of YOLO. The frame extraction process is done in such a way that it reduces the time complexity by eliminating the frames which don't contain vehicles. The quality of the detected license plates is improved using the Image Quality Enhancer. Image Transformer is used to tackle the problem of recognization of license plates when they are at an angle. The Smart Recognizer consists of two parts again: Tesseract-OCR, Error Decoder. The Tesseract-OCR is used to recognize the license number and the characters that are wrongly detected by the Tesseract-OCR are corrected by Error Detector.

\begin{figure}[h!]
  
  \centering
  \includegraphics[width=0.5\textwidth]{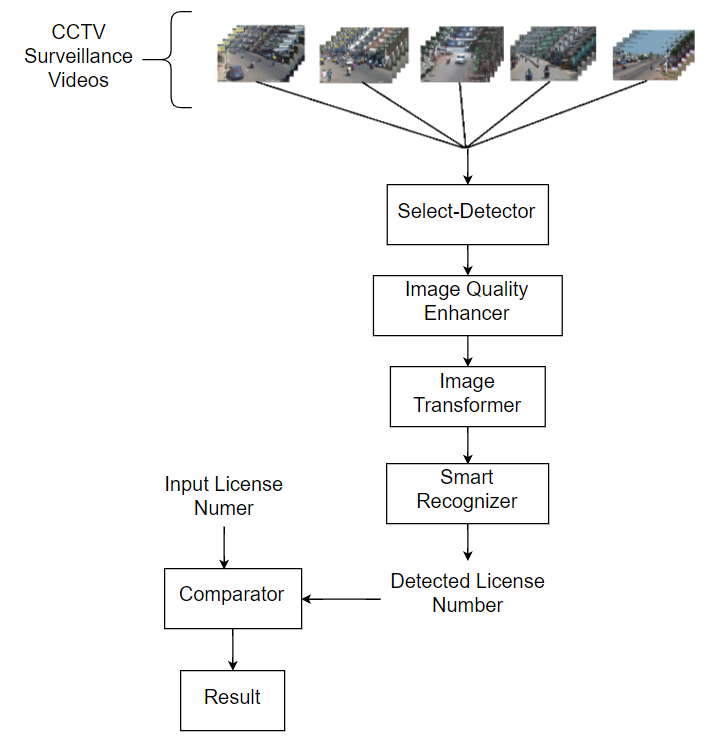}
  \caption{Overview Of Proposed Framework}
  \label{fig:1} 
\end{figure}

\subsection{Select-Detector}

\begin{figure}[h!]
  
  \centering
  \includegraphics[width=0.5\textwidth]{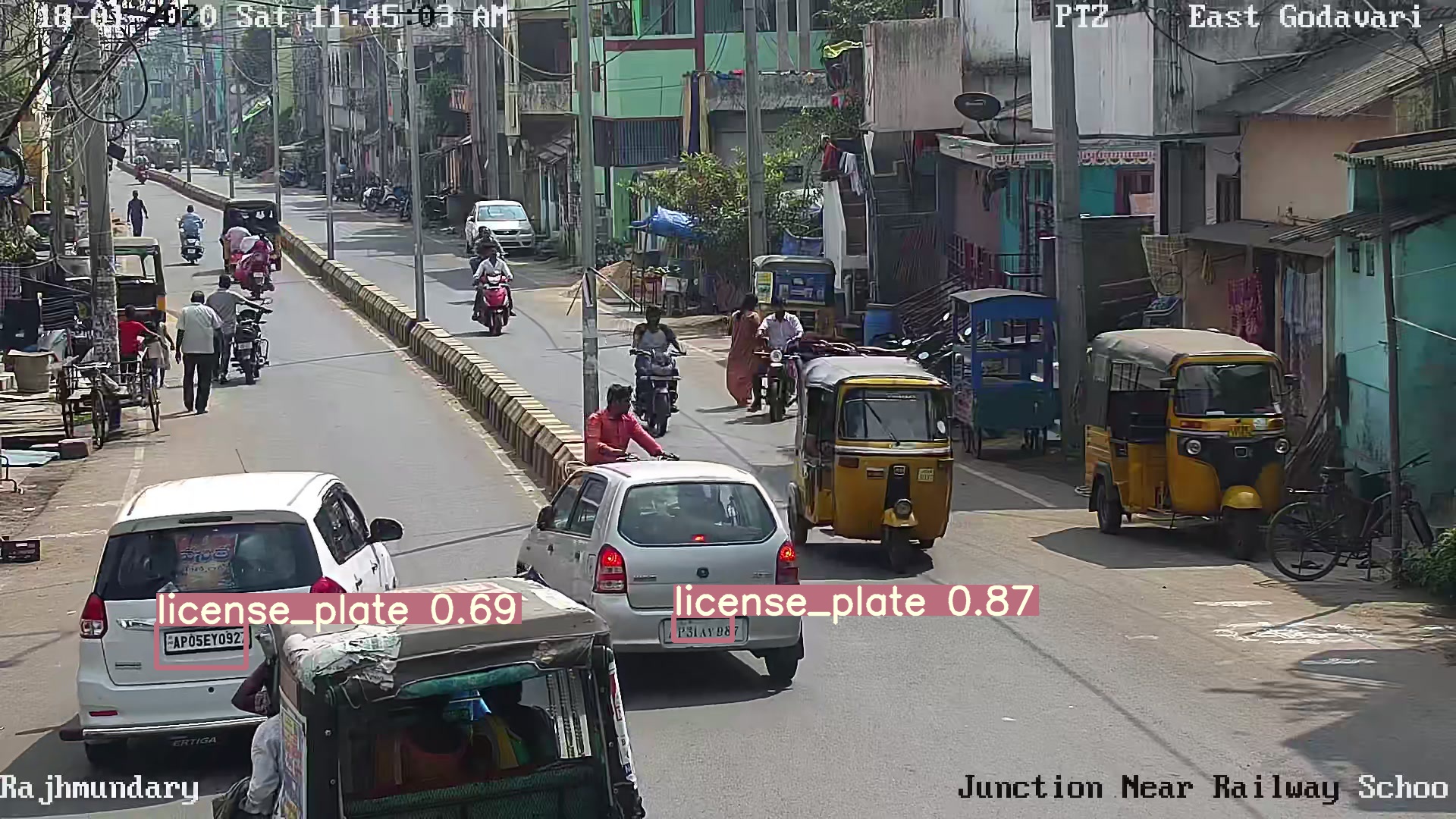}
  \caption{Car's License Plate Detection by YOLOv3}
\end{figure}

Select-Detector consists of a YOLOv3 object detector trained to detect the license plates of cars in an image. It is also used for the effective extraction of frames from CCTV surveillance videos. Figure 2, shows the license plate detection by YOLOv3. YOLO gives the coordinates of the region of interest(ROI) in the input frame where the license plate is located. The boarders of the ROI are extended further by 20 pixels on the right and left side. Which is done to make sure that all the characters in the license plate are considered and this is depicted in Figure 3. 

\begin{figure}[h!]
  
  \centering
  \includegraphics[width=0.5\textwidth]{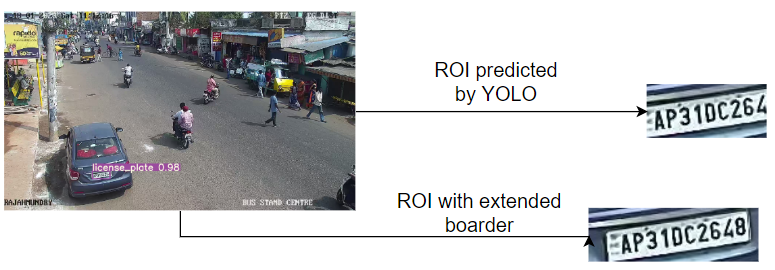}
  \caption{ROI predicted by YOLOv3 with extended border}
\end{figure}

\begin{figure*}
  \includegraphics[width=\textwidth]{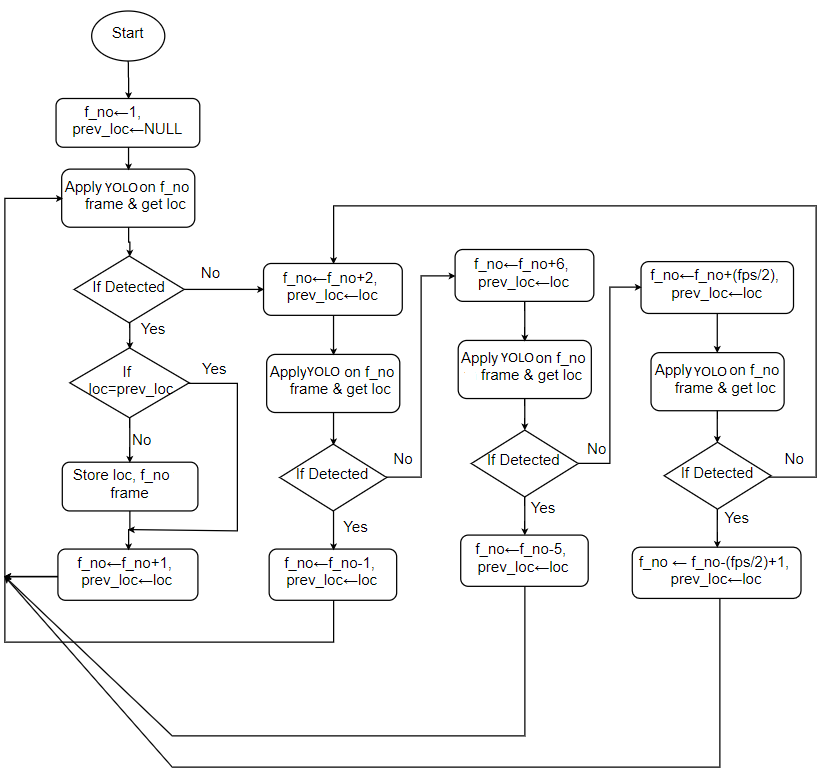}
  \caption{Algorithm for selecting only required frames from videos for reducing the time complexity.f\_no represents the frame number, loc represents the location coordinates of license plate predicted by YOLO, prev\_loc is a variable to store the previous location, fps denotes the frames per second.}
\end{figure*}

Generally in any vehicle tracking system, the license plate detection algorithm is applied on every single frame which may increase the time of computation. So, it is necessary to select only those frames which contain vehicles instead of selecting every frame. There is also a case when the vehicle is at rest for some time, but several frames are extracted during this time of rest, which may also increase the processing time in later stage. Hence, it is important to identify the vehicles that are at rest and extracting only one frame in the steady state. The algorithm for extracting frames efficiently is depicted in Figure 4. This algorithm for selecting only required frames works efficiently by assuming that, any vehicle once appeared in the field of sight of the camera, stays in the field of view for the next half a second and is detected by the object detector. Considering frames per second(fps) equal to 25, every vehicle should appear in at least 13 consecutive frames.

\subsection{Image Quality Enhancer}

To solve the problems such as noise, low contrast, shadow, etc in the license plate image, image quality enhancement technique is applied.
Pix2pix GAN\cite{b12} is used for this purpose. It is a model designed for general purpose image-to-image translation.
Pix2pix learns a function to map from a low-quality input image to a high-quality output image using a conditional generative adversarial network (cGAN) . The network consists of two main parts, the Generator, and the Discriminator. The Generator transforms the low-quality input image to get the high quality output image. The Discriminator estimates the similarity of the input image to an unknown image (either a target image from the dataset or an output image from the generator) and tries to guess if this was the actual image produced by the generator.
The Generator comprises of an encoder-decoder like structure with skip-connections giving it a U-net shaped architecture as shown in Figure 5.

The Discriminator takes in two images, an input image and an unknown image (which will be either a target or output image from the generator), and tries to decide if the other image was produced by the generator or not.
The discriminator model consist of a sequence of standard Convolution, Batch Normalization, Rectified Linear Unit(ReLU) blocks similar to deep convolutional neural networks as shown in Figure 6. All the ReLUs in the encoder of the generator and the discriminator are leaky, whereas the  ReLUs in the decoder of the generator are not leaky. Pix2pix GAN is trained to generate a new high-quality license plate image when given a low-quality license plate image with low-resolution, low-contrast, shadow, noise etc.

\begin{figure}[h!]
  
  \centering
  \includegraphics[width=0.5\textwidth]{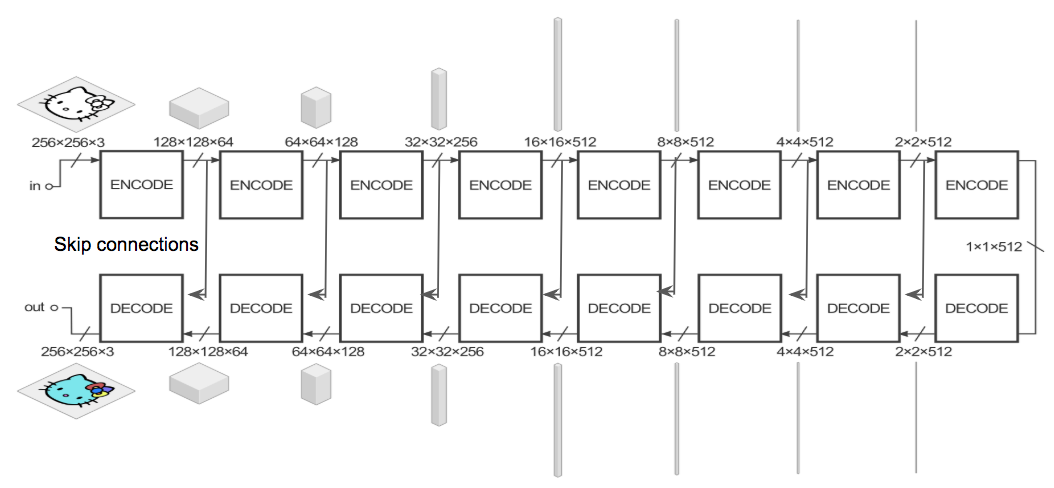}
  \caption{Pix2pix's Generator architecture
  (Source: https://neurohive.io/en/popular-networks/pix2pix-image-to-image-translation/)}
\end{figure}

\begin{figure}[h!]
  
  \centering
  \includegraphics[width=0.5\textwidth]{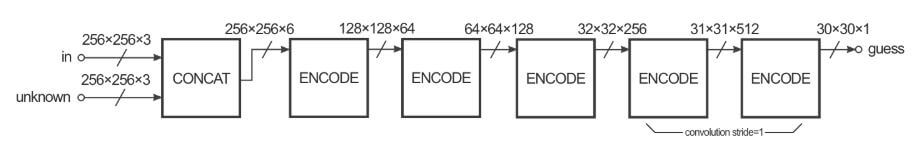}
  \caption{Pix2pix's Discriminator architecture  (Source:https://neurohive.io/en/popular-networks/pix2pix-image-to-image-translation/)}
\end{figure}
\subsection{Image Transformer}

The high-quality image generated by the Image Quality Enhancer is taken as an input to the Image Transformer. The function of this component is to solve the problem of wrong recognition when the license plates are not horizontal(i.e at an angle). It outputs a set of images that are obtained by transforming the input image to different rotation and cropping levels, assuming that at least one of these images would be closer to the horizontal level and that can give better recognition results compared to the input image which was not horizontal. Then the recognizer is applied to each of these sets of images and the results are noted. During testing, distance is calculated between the suspicious car's license number and the recognition result of every image obtained from Image Transformer. Based on a threshold for distance, the input image is classified whether it matches with the suspicious car's  number or not. Rotating an image is done in the range of 12 degree clockwise to 12 degree anti-clockwise while cropping is done by reducing the border from 0 to 25 pixels. Rotation and cropping are done at 20 different levels to get 20 different sub-images for each image, in a hope that one of the sub-images is approximately horizontal. 
\begin{figure}[h!]
  
  \centering
  \includegraphics[width=0.5\textwidth]{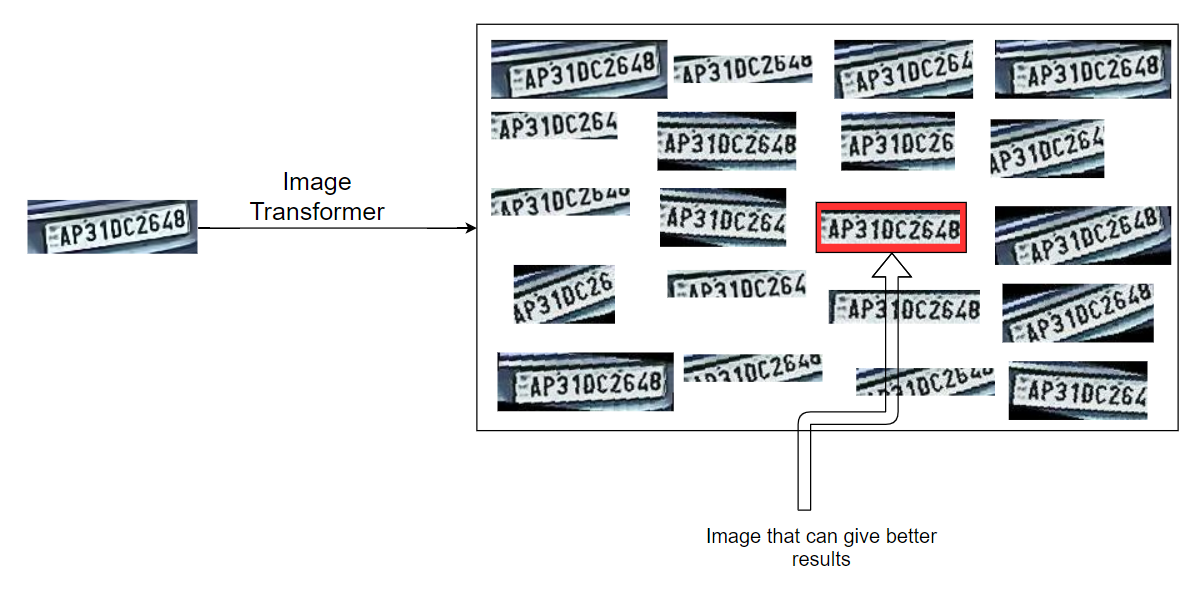}
  \caption{Image Transformer}
\end{figure}

\subsection{Smart Recognizer}

The final step of the framework is to recognize the license plates generated by the Image Transformer. Smart Recognizer uses Tesseract-OCR\cite{b13} is used to perform this task.The accuracy of the Tesseract is fairly high and can be improved significantly with a well-designed image preprocessing pipeline. The procedure behind preprocessing involves OpenCV's median filter, contrast enhancer and BGR2grayscale conversion. The primary character classifier function in Tesseract OCR is based on an implementation of a Long Short-Term Memory neural network or LSTM network. LSTM neural networks outperform all other alternative neural network architecture models for this type of pattern recognition.
Though Tesseract works well at recognizing most of the license plates, in some cases it fails to predict some of the characters in the plate correctly, where the characters look similar, like zero in numbers and 'O' in alphabets. 
Thus,an algorithm called Error-Detector, is developed using the standard format of Indian license plates and by identifying the characters and numbers which look similar to each other to increase the accuracy of recognition even more.

The current format of the vehicle registration plate (commonly known as number plate) number, issued by the district-level RTO consists of 4 parts, They are:
\begin{itemize}
\item The first two characters of the license plate are the alphabets that represent the Union Territory or State to which the vehicle is registered. e.g., TS 09 UB 8902
\item The next two characters are the numbers that represent the sequential number of a district. e.g. KA 51 MD 4182
\item The third part consists of one, two or three alphabets or no alphabets at all. This represents the ongoing series of an RTO and/or vehicle classification.
\item The fourth part is a number from 1 to 9999, unique to each plate. An alphabet is prefixed when the 4 digit number runs out and then two alphabets and so on.
\item The fifth part is an international oval "IND" and the above it a hologram having a Chakra. However, not all plates have these features.
\end{itemize}

The algorithm works on the principle that whenever a number is recognized in the place where the alphabet should be present according to the format of the Indian number license plate, the number is replaced with its alphabetic analog, and whenever an alphabet is recognized in the place where a number should be present, the alphabet is replaced with its numeric analog.  The numeric-alphabetic analogs used, are listed in the Fig. 8 and the flowchart for Error-Detector is depicted in Fig. 9

\begin{figure}[h!]
  
  \centering
  \includegraphics[width=0.5\textwidth]{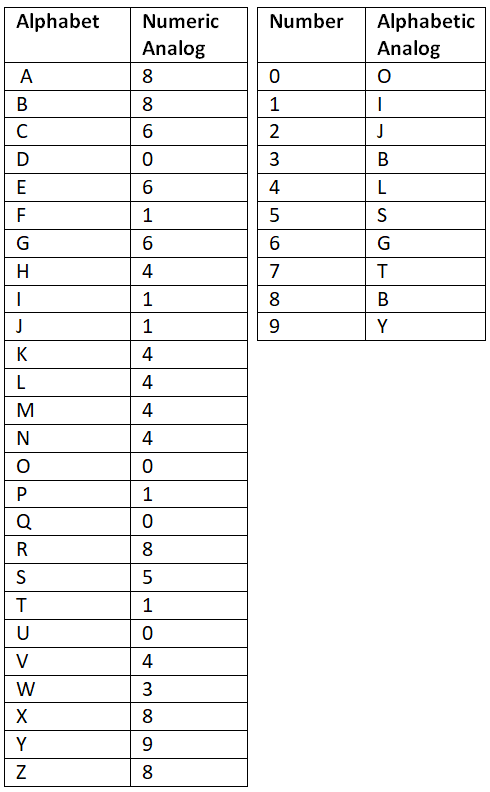}
  \caption{Numeric-Alphabetic analogs}
\end{figure}

\begin{figure*}[h!]
  
  \centering
  \includegraphics[width=\textwidth]{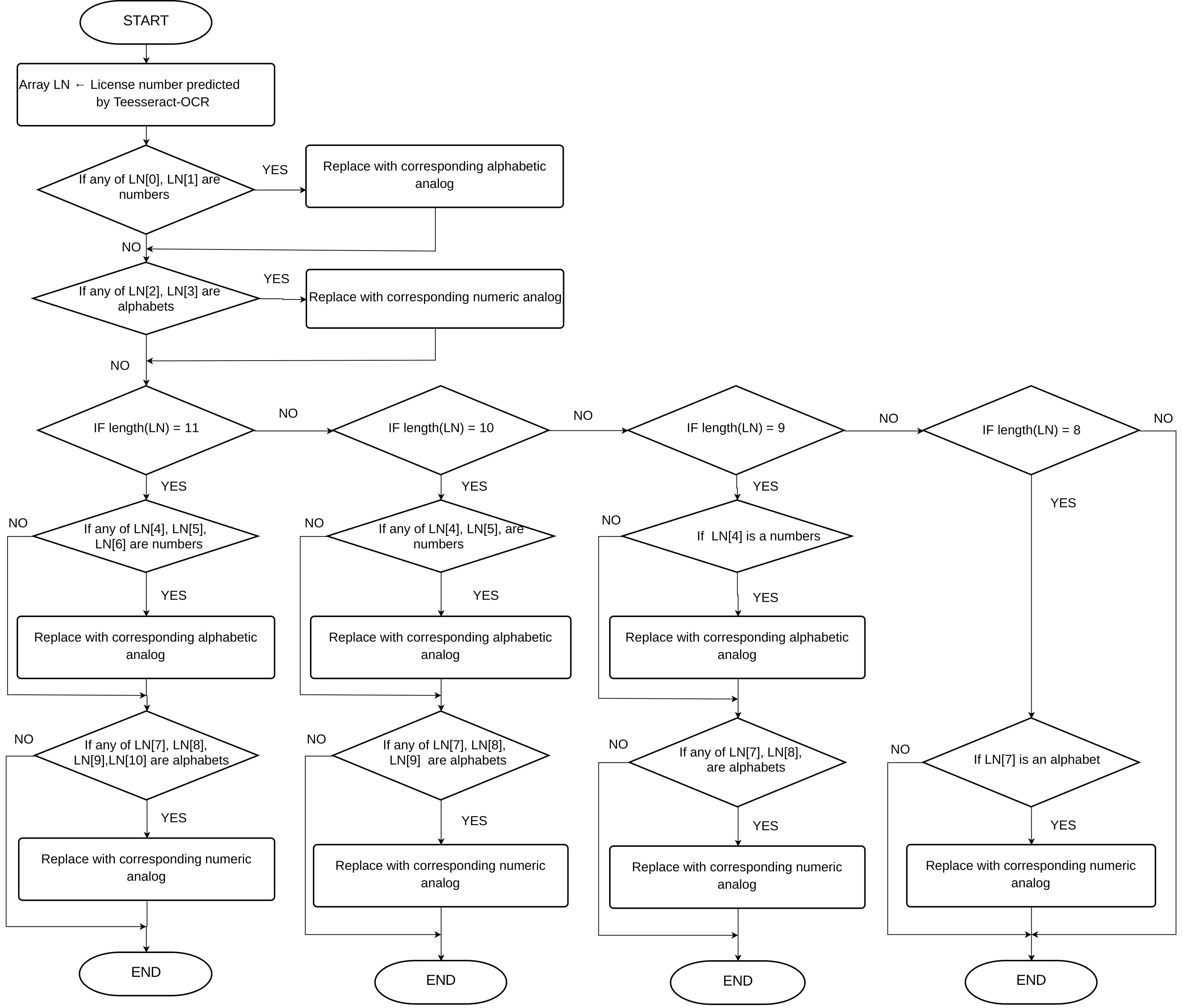}
  \caption{Error-Detector}
\end{figure*}

\subsection{Comparator}

The comparator takes in two license numbers and tries to find the distance between these two numbers. One of the two numbers is the suspicious car's license number and the other is the predicted one generated by the smart recognizer. The distance here is defined as the number of characters of the two license numbers that are mismatched. Based on the threshold value for the distance, the license number identified by the framework is classified as matched or mismatched. If the detected car's license number matches with the required suspicious car's license number, the time and the location of the video in which the car is detected are recorded. For the proposed framework threshold value for the distance of mismatch is taken as 2.

\section{Experiments And Results}

\subsection{Dataset Acquisition }
The proposed framework is evaluated on a set of Government's CCTV road surveillance videos obtained from the Urban Police District, Rajamahendravaram,Andhra Pradesh,India. The cameras used to capture footage are not dedicatedly selected or set up for automatic license plate detection and recognition. The dataset contains 5 videos of varying time-duration obtained from the CCTV cameras set up by the Government in different areas in the city.The resolution of each video is  1920 x 1080. Due to improper functioning of cameras, videos had abruptly paused for few seconds. This was resolved by removing the paused frames from the video. For example if we consider a video of 1 minute, and the system got paused in between 20-30 seconds time interval, this portion is then removed and the remaining video that was recorded is added to the preceding portion i.e after $19^{th}$ second. 
Re-Identification(ReID) license plates dataset provided by Jakubet et al.\cite{b14} is used for training of Image Quality Enhancer. It contains 182,336 color license plate images of different lengths, image blur and slight occlusion.

\subsection{Image Quality Enhancer}
A total of 326 high quality images are selected from the ReID dataset and are made noisy, blur, low contrast using python's OpenCV. As the images were of different sizes, all of them were resized to 256x256. Random Gaussian noise was added to images.
The images were blurred using Gaussian Blur, Median Blur, Bilateral Blur randomly. The dataset thus prepared, contains 306 original-low quality image pairs for training and 20 image pairs for validation. The training was done for 40 epochs on a batch size of 64. The training process took about less than an hour to complete when trained on NVIDIA Tesla K80 GPU.
The pix2pix network and its weights obtained after training are saved and are used as an image quality enhancer for testing. The results shown by this trained model on our dataset are presented in Figure 10.

\begin{figure}[h!]
  
  \centering
  \includegraphics[width=0.5\textwidth]{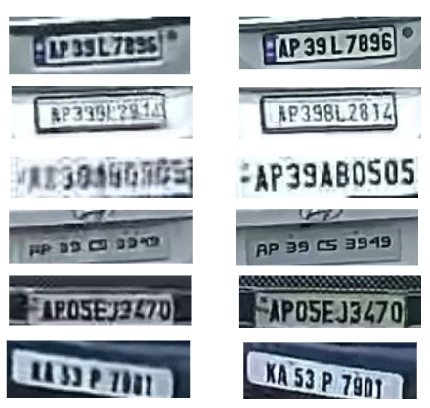}
  \caption{Image Quality Enhancer- left side images shows the low quality images and right side images shows the high quality images generated by Image Quality Enhancer}
\end{figure}

The performance of the proposed system is evaluated on the acquired dataset. 
The license numbers of the cars that are clearly visible from the dataset are manually taken as a test set for evaluation. The test set taken contains 53 license numbers. The bar graph in Figure 11, shows the percentage of cars that are correctly detected, incorrectly detected and not detected respectively. The testing accuracy achieved by the proposed model for detecting the license number was 87\%.
During the evaluation, all these set of license numbers are passed one by one to the proposed framework and the framework predicts the location and the time when a specific license number is seen in the dataset. The results predicted by the proposed model are shown in the Figure 12. 

\begin{figure}[h!]
  
  \centering
  \includegraphics[width=0.5\textwidth]{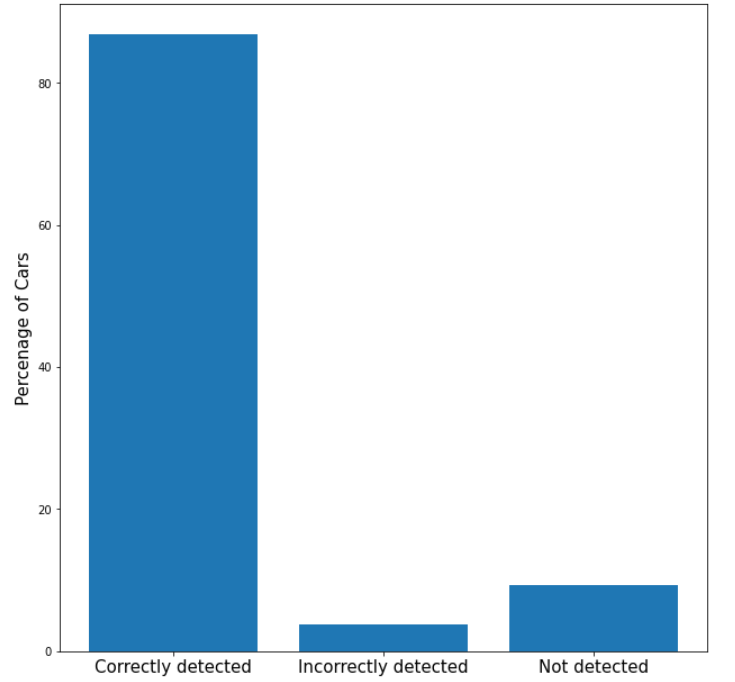}
  \caption{Bar graph showing percentage of cars that are correctly detected, incorrectly detected, not detected}
\end{figure}

\begin{figure*}
  
  \centering
  \includegraphics[width=0.7\textwidth]{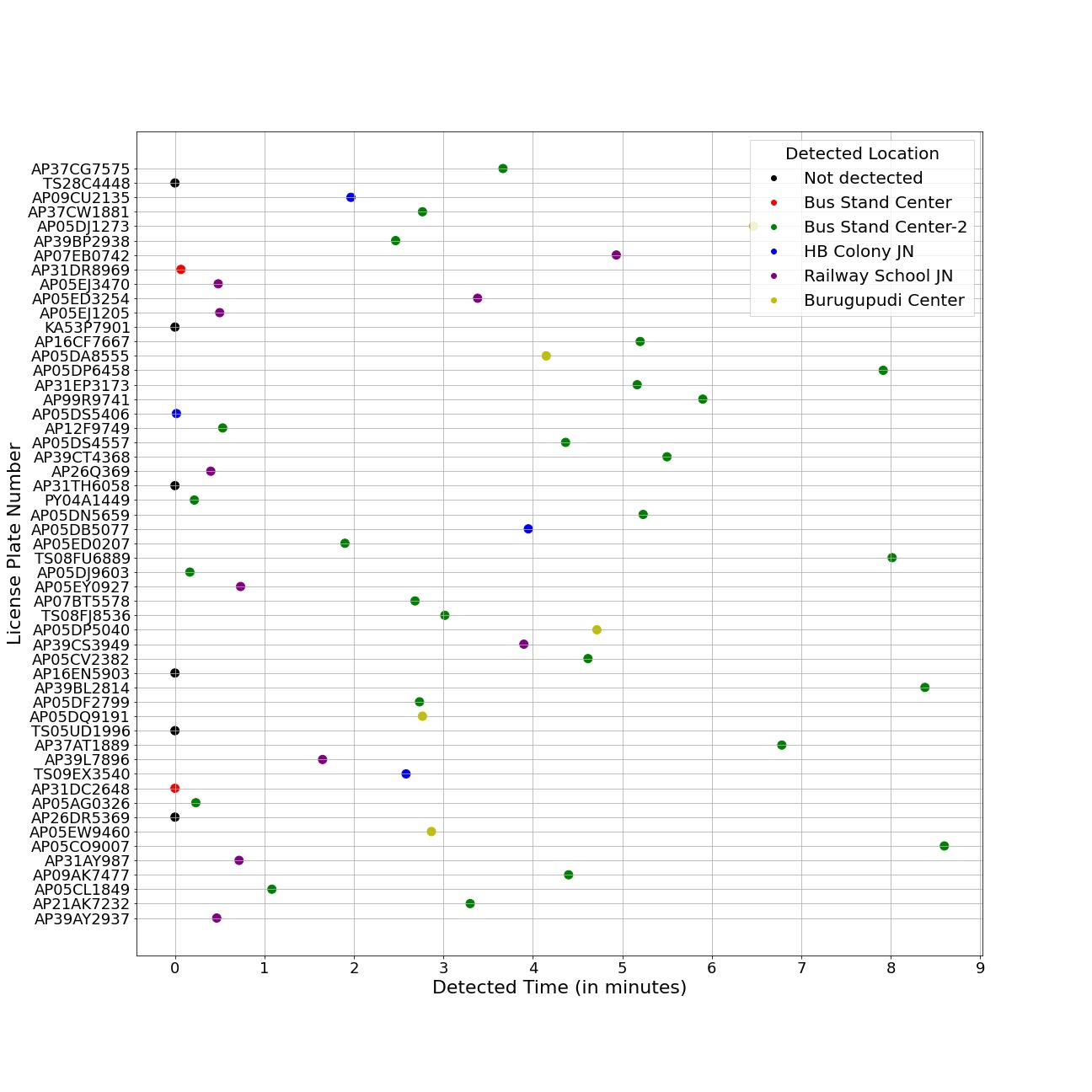}
  \caption{Results predicted by the proposed framework. The scatter points gives the predicted location when and where the car was last seen.}
\end{figure*}

The Y-axis represents the license numbers, the X-axis shows the predicted time and the color of the scatter points shows the predicted location when and where the car with the given license number was spotted.

\section{Conclusion}
An intelligent deep learning based system is implemented to detect the suspicious or missing cars using their license numbers from a set of CCTV videos captured from different locations. The system has achieved impressive results in detecting the cars with an accuracy of 87\%.
With this automated process, no human effort will be required to manually verify each and every car from the CCTV videos. This system
can be used by the Police Department for carrying out their investigation fast without human intervention in cases where there is a requirement for tracking the location of a car when it enters the city traffic, with the help of CCTVs fixed at different locations in the city. The give model can perform well, when compared to other tracking systems available.
In the future, the proposed system can be used for traffic management and identification of vehicles violating traffic rules by replacing an intelligent auto-rotate model instead of the Image Transformer used in this paper.


%
 \section*{Conflict of interest}
The authors declare that they have no conflict of interest.



\end{document}